\documentclass[11pt]{article}

\usepackage[margin=1in]{geometry}
\usepackage{times}
\usepackage[utf8]{inputenc}
\usepackage[T1]{fontenc}
\usepackage{microtype}
\usepackage{booktabs}
\usepackage{amsmath}
\usepackage{amssymb}
\usepackage{graphicx}
\usepackage{multirow}
\usepackage{array}
\usepackage{url}
\usepackage[numbers,sort&compress]{natbib}
\usepackage{pgfplots}
\pgfplotsset{compat=1.18}
\usepackage[hidelinks]{hyperref}
\usepackage{authblk}

\title{Attention Degradation, Function Token Anchoring, and the Limits
of Attention-Based Intervention in Large Language Models}

\author[1]{Sagar Dangal}
\author[2]{Manoj Shakya}
\affil[1]{London Metropolitan University / Islington College, Kathmandu, Nepal\\ \texttt{ragaslagnad28@gmail.com}}
\affil[2]{Department of Computer Science and Engineering, Kathmandu University, Dhulikhel, Nepal}

\date{}

\begin{document}
\maketitle

\begin{abstract}
Mean cross-positional attention degradation is widely reported in
transformer interpretability, yet whether it causally limits
contextual retrieval remains untested. We present six coordinated
experiments across GPT-2, LLaMA-3.2-1B/3B, OPT-1.3B, and distilgpt2.
We first characterise short-term (5--100 token) attention degradation,
finding a universal exponential-then-plateau pattern whose rate is
inversely correlated with depth, with distinct layer-wise entropy
signatures per architecture. Function token anchoring proves
architecture-dependent: OPT-1.3B (absolute positional encoding) shows
distance-dependent preposition specificity, GPT-2 shows uniform
non-specific dependence, and LLaMA (RoPE) shows reversal at long
distances. Strategic comma insertion at clause boundaries causally
reduces prediction degradation in the 40--80 token range, with the
benefit tied to syntactic boundary alignment rather than token
density. We then test the mechanism causally: Relay-Aware Attention
(RAA), which biases attention logits toward function token positions,
verifiably increases attention mass by 16--24\% yet yields null
effects on GPT-2 and LLaMA-1B, preliminary harm on LLaMA-3B, and a
mixed effect on OPT-1.3B that nets to approximately zero. Multi-fact
retrieval probes further show that degradation rate does not predict
retrieval accuracy across models. We conclude that mean attention
degradation is largely descriptive rather than prescriptive: function
tokens contribute through what their hidden states compute, not
through the attention they receive --- with implications for
interpretability methodology and attention-score-based inference
optimisations such as KV-cache eviction.
\end{abstract}


\section{Introduction}
\label{sec:intro}

The self-attention mechanism in transformer-based language models
\citep{vaswani2017attention} allows each token to attend to every
preceding position, weighted by learned query-key similarity. In
principle this enables arbitrary long-range dependency modelling; in
practice, large language models (LLMs) exhibit systematic limitations
in utilising contextual information as input length increases.

The ``Lost in the Middle'' phenomenon \citep{liu2024lost} demonstrates
that information placed in the middle of long contexts is retrieved
substantially less accurately than information at the beginning or
end---a primacy-recency bias observed across models with context
windows from 4K to 128K tokens. \citet{hong2025context} documented
``context rot'' across 18 frontier models, with performance degrading
well before nominal context limits. \citet{yang2025ikod} identified
analogous fading focus patterns in vision-language models. These
studies converge on the conclusion that transformer attention does not
distribute uniformly across input.

Short-term degradation---over spans of 5 to 100 tokens---has been
comparatively underexplored, yet this range corresponds to the typical
span of a clause or short sentence: the fundamental unit of syntactic
composition. \citet{zhang2025function} proposed the \emph{function
token hypothesis}: that articles, prepositions, and punctuation marks
activate predictive features from context during inference. We develop
a complementary \emph{relay-chain} interpretation: if each function
token has a bounded radius of contextual influence, the high density
of function tokens in natural prose may produce a chain of overlapping
anchors supporting clause-length context retention.

Three questions motivate this work: (1)~How does short-term attention
degradation vary across architectures with different positional
encoding schemes and depths? (2)~Do function tokens act as structural
anchors for contextual retention, and is this role
architecture-dependent? (3)~Is mean attention degradation a causal
bottleneck limiting retrieval, or merely a descriptive correlate of
computations that occur elsewhere?

The first two questions are addressed through four correlational and
interventional experiments (Experiments~1--4): we characterise the
degradation profile, probe function token specificity via cloze
substitution, test causal benefit via strategic token insertion, and
verify corpus-level relay-chain coverage. The third question is
addressed through two causal experiments (Experiments~5--6): we
operationalise the relay chain as an attention-logit bias toward
function token positions (Relay-Aware Attention, RAA) and test whether
verified attention redistribution improves behaviour, and we test
whether degradation rate predicts multi-fact retrieval performance
across models.

The answer to the third question is decisively negative, and
constitutes our central methodological finding: \textbf{mean
cross-positional attention degradation is largely descriptive rather
than prescriptive}. A verified 16--24\% increase in attention mass at
function token positions produces null effects on GPT-2 and LLaMA-1B,
preliminary evidence of harm on LLaMA-3B, and a distance-dependent
mixed effect on OPT-1.3B that nets to approximately zero; and
degradation rate fails to predict retrieval accuracy across models.
Where function tokens contribute to contextual processing---and
Experiments~2--3 show they genuinely do, in architecture-specific
ways---they do so through the contextual information accumulated in
their hidden states, not through the attention routing that mean
attention analysis captures.

\subsection{Research Questions}

\begin{description}
\item[RQ1] How does attention strength degrade with token distance in
  the 5--100 token range, and does the rate vary across architectures?
\item[RQ2] Do function tokens specifically contribute to contextual
  memory retention, or is replacement damage comparable to removing
  any token?
\item[RQ3] Can inserting function tokens at strategic positions reduce
  prediction degradation, specifically for function tokens over
  matched controls?
\item[RQ4] Are function tokens dense enough in natural text for their
  effective radii to form a continuous relay chain?
\item[RQ5] Can the relay-chain mechanism be operationalised via
  attention-logit biasing toward function token positions, and does
  this improve contextual retrieval?
\item[RQ6] Does mean attention degradation rate predict contextual
  retrieval performance across architectures?
\end{description}

\subsection{Contributions}

\begin{enumerate}
\item The first systematic cross-architecture characterisation of
  short-term attention degradation (5--100 tokens) across four
  architecturally diverse models, establishing an
  exponential-then-plateau pattern universal in shape but
  architecture-dependent in rate, together with layer-wise entropy
  signatures that explain the cross-model differences.
\item First systematic evidence that function token specificity is
  architecture-dependent: OPT-1.3B shows distance-dependent
  preposition specificity; GPT-2 shows uniform non-specific
  dependence; LLaMA shows reversal at long distances consistent with
  RoPE redundancy.
\item Causal evidence that comma insertion at clause boundaries
  reduces prediction degradation (40--80 tokens), specific to function
  tokens over matched content-token controls, including rank-based
  evidence in LLaMA-3.2-3B and a comparison of four insertion
  strategies showing that syntactic boundary alignment, not token
  density, drives the benefit.
\item Corpus-level confirmation that function token density in natural
  English provides 83--89\% positional coverage per category
  (96--98\% combined), stable across three tokenizers.
\item A direct causal test of the relay-chain hypothesis via
  Relay-Aware Attention (RAA), demonstrating that biasing attention
  toward function token positions produces null, negative, or marginal
  effects despite verified attention mass redistribution.
\item Multi-fact retrieval probes demonstrating that mean degradation
  rate does not predict retrieval accuracy: model capacity, not
  degradation rate, is the dividing line.
\item The methodological conclusion that mean cross-positional
  attention is descriptive rather than prescriptive, extending the
  attention-as-explanation debate to autoregressive LMs with a
  verified interventional design, with practical implications for
  attention-score-based inference optimisations such as KV-cache
  eviction.
\end{enumerate}

\section{Related Work}
\label{sec:related}

\subsection{Attention Mechanisms and Positional Encoding}

The transformer \citep{vaswani2017attention} employs multi-head scaled
dot-product attention. \citet{clark2019bert} showed BERT heads develop
distinct specialisations; \citet{voita2019analyzing} showed most heads
are prunable. GPT-2 \citep{radford2019language} and OPT
\citep{zhang2022opt} use learned absolute positional embeddings; LLaMA
\citep{dubey2024llama} uses Rotary Position Embedding (RoPE;
\citealt{su2024roformer}), encoding relative position directly into
the attention computation. LLaMA-3.2 additionally uses Grouped-Query
Attention (GQA); we treat query-level attention weights as the unit of
analysis, consistent with prior interpretability work.

\subsection{Attention Degradation}

\citet{liu2024lost} demonstrated U-shaped performance curves on
multi-document QA tasks at the 4K--128K token scale.
\citet{hong2025context} documented systematic degradation across 18
frontier LLMs. \citet{yang2025ikod} extended findings to
vision-language models. \citet{gupta2025depth} found LLMs use depth
non-uniformly, with function words among the earliest correctly
predicted. None of this prior work examines the 5--100 token
clause-level range, which is the regime addressed here.

\subsection{Attention as Explanation: Descriptive vs.\ Causal}
\label{sec:attn-explanation}

Whether attention weights constitute mechanistic evidence of model
behaviour is contested. \citet{jain2019attention} showed that
attention distributions are not faithful explanations in encoder
classifiers: perturbing attention often has minimal effect on
predictions. \citet{wiegreffe2019attention} argued that this does not
fully refute an explanatory role, and the debate has continued
\citep{bibal2022attention}. These studies were largely conducted on
encoder models in classification settings, relying primarily on
sensitivity analysis. Our work extends this question to autoregressive
LMs and contributes a direct interventional test: we verify that
attention mass is redistributed (16--24\%), then ask whether behaviour
changes in the predicted direction. The interventional framing
addresses the key limitation of correlation-based faithfulness
analyses.

\subsection{The Function Token Hypothesis}

\citet{xiao2023streamingllm} identified the attention sink: a
disproportionate fraction of attention mass allocated to the first few
tokens regardless of semantic relevance. \citet{zhang2025function}
proposed that function tokens activate predictive features from
context and drive next-token prediction. We extend this into a
relay-chain interpretation and, critically, test it causally.
\citet{qian2025demystifying} find thinking tokens in chain-of-thought
correspond to mutual-information peaks with the final answer---an
analogous but distinct category from our syntactic function tokens
(see also \citealt{ding2025thinking}). \citet{zhang2025rft} showed
reinforced functional token tuning improves multi-step reasoning
without architectural modification.

\subsection{Attention Modification and Attention-Based Selection}

\citet{press2022alibi} introduced ALiBi, subtracting linear distance
penalties from attention logits to enable length extrapolation. Our
RAA inverts this logic, rewarding structural positions rather than
penalising distance. \citet{meng2022locating} showed factual
associations in GPT are primarily localised in MLP layers, not
attention heads---motivating our hypothesis that mean attention
degradation may not capture the mechanisms responsible for retrieval.
A separate line of work uses accumulated attention scores to select
which tokens to retain during inference, including H2O
\citep{zhang2023h2o} and SnapKV \citep{li2024snapkv}; these methods
implicitly assume that attention mass tracks causal importance, an
assumption our experiments test directly.

\section{Methodology}
\label{sec:method}

\subsection{Models}

\begin{table}[t]
\centering
\small
\begin{tabular}{lrrrrl}
\toprule
Model & Params & L & QH & KVH & Pos.\ Enc. \\
\midrule
GPT-2        & 124M  & 12 & 12 & 12 & Absolute \\
LLaMA-3.2-1B & 1.24B & 16 & 32 &  8 & RoPE \\
LLaMA-3.2-3B & 3.21B & 28 & 24 &  8 & RoPE \\
OPT-1.3B     & 1.32B & 24 & 32 & 32 & Absolute \\
\bottomrule
\end{tabular}
\caption{Models used across experiments. L=layers, QH=query heads,
KVH=KV heads. All are base pretrained models. distilgpt2 (6 layers,
82M) is additionally used in Experiment~6. LLaMA-3.2 uses GQA;
attention is analysed at the query-head level.}
\label{tab:models}
\end{table}

Table~\ref{tab:models} summarises the models. The set spans a
26$\times$ parameter range (124M--3.21B), a 2.3$\times$ depth range
(12--28 layers), and both major positional encoding families (learned
absolute embeddings vs.\ RoPE), enabling the architecture-comparative
analysis that is central to this work.

\subsection{Datasets}

WikiText-103-raw-v1 \citep{merity2017pointer} was used for
Experiments~1 and~5; WikiText-2-raw-v1 for Experiments~2, 3, and~4.
Texts shorter than 50 characters were filtered; all texts were
tokenised with each model's native tokenizer and truncated to 512
tokens. 100 samples were drawn from the test split, with 5--10 random
seeds per model. Experiment~6 uses a synthetic multi-fact interference
design (\S\ref{sec:exp6-method}).

\subsection{Function Token Categories}

Following \citet{zhang2025function}: \textbf{Articles} (\emph{the, a,
an}); \textbf{Prepositions} (\emph{of, to, in, for, on, at, by, with,
from, as}); \textbf{Punctuation} (\texttt{,}~\texttt{.}~\texttt{;}~%
\texttt{:}~\texttt{!}~\texttt{?}). Categories are operationalised at
the token level using each model's native tokenizer.

\subsection{Experiment 1: Baseline Attention Degradation}
\label{sec:exp1-method}

A full forward pass was performed with attention outputs retained.
For each of 30 randomly sampled query positions per text and each
target distance $d \in \{5,10,\ldots,100\}$, the attention score is
the mean across all layers and heads of the attention weight from
position $q$ to $q-d$. Degradation at distance $d$ is defined as
\begin{equation}
  \text{Degradation}(d) =
  \left(1 - \frac{\text{Attn}(d)}{\text{Attn}(5)}\right) \times 100\%.
\end{equation}
Statistical significance was assessed via paired $t$-tests across 10
seed-level means. Measurements were taken at $d \in
\{5,10,\ldots,95\}$; $d=100$ was excluded due to an attention-sink
boundary artefact at exactly this distance from position~0 in
512-token sequences (\S\ref{sec:d100}). Layer-wise attention entropy
(Shannon entropy of the attention distribution over prior positions,
averaged across query positions and texts) was additionally computed
for each model.

\subsection{Experiment 2: Cloze Substitution}
\label{sec:exp2-method}

A content word is selected as prediction target; baseline probability
and rank are measured. For each distance $d$ and function token
category $c$, all function tokens of category $c$ within the window
$[\text{target}-d,\,\text{target}]$ are replaced with a neutral
substitution token (a newline character).\footnote{Replacing function
tokens with a newline introduces a potential confound, as newline is
itself structurally meaningful in some tokenizers. A rare
out-of-vocabulary token would be a cleaner substitute; we treat this
as a limitation (\S\ref{sec:limitations}).} A matched control replaces
an equal number of randomly selected content tokens. Specificity is
defined as function token damage systematically exceeding control
damage at a given distance. Bonferroni correction was applied across
$\approx$63 distance$\times$category tests.

\paragraph{Inter-run stability.} Repeat runs of Experiment~2 with
identical configuration produced mean function-token replacement
$\Delta p$ estimates varying by up to $2\times$ across runs,
suggesting results in the 5--100 sample range are sensitive to sample
composition. Specificity percentages should be interpreted with this
variance in mind.

\subsection{Experiment 3: Function Token Insertion}
\label{sec:exp3-method}

Four insertion strategies were tested: (1)~\emph{Boundary}: inserts a
comma after content words directly followed by other content words,
approximating natural clause-boundary positions; (2)~\emph{Gap-fill}:
inserts at positions where no function token exists within the
effective radius; (3)~\emph{Interval-10} and (4)~\emph{Interval-20}:
inserts every 10 or 20 tokens regardless of syntactic structure. A
matched control inserts an equal number of random content tokens at
random positions. Significance was assessed via paired $t$-tests on
per-seed mean probability and rank changes. Because the control
inserts at random positions rather than at the positions used by the
function-token strategies, observed differences partially reflect
insertion position as well as token identity; we treat this as a
limitation (\S\ref{sec:limitations}).

\subsection{Experiment 4: Relay Chain Coverage}
\label{sec:exp4-method}

For each text and function token category, gaps between consecutive
function tokens are measured. Coverage is the percentage of
non-function-token positions falling within the effective radius of at
least one function token. Effective radii are derived empirically from
Experiment~2 results (articles: 60 tokens; prepositions: 25;
punctuation: 35). Note that article and punctuation radii are derived
from conditions showing weak or absent specificity in most models
(\S\ref{sec:exp2-results}); coverage figures for these categories thus
primarily reflect corpus density rather than model-specific anchoring,
and should be interpreted accordingly. Three tokenizers (GPT-2 BPE,
LLaMA SentencePiece, OPT BPE) are compared.

\subsection{Experiment 5: Relay-Aware Attention (RAA)}
\label{sec:exp5-method}

Standard scaled dot-product attention is
\begin{equation}
  A(q,j) = \operatorname{softmax}\!\left(
    \frac{QK^\top}{\sqrt{d_k}}\right)[q,j].
\end{equation}
RAA adds a position-dependent bias before the softmax:
\begin{equation}
  A'(q,j) = \operatorname{softmax}\!\left(
    \frac{QK^\top}{\sqrt{d_k}} + B(j)\right)[q,j],
\end{equation}
where $B(j)=\delta$ if token $j$ is a function token and $B(j)=0$
otherwise. The scalar $\delta$ is tested at $\{0.0, 0.1, 0.5, 1.0,
2.0\}$, applied uniformly across all layers and heads. RAA was tested
on GPT-2 (10 seeds, 100 samples), OPT-1.3B (5 seeds, 50 samples),
LLaMA-1B (2 seeds), and LLaMA-3B (2 seeds).

\paragraph{Attention verification.} Before interpreting effect sizes,
we verify that RAA materially changes attention distributions. At
$\delta=2.0$, mean attention at function-token positions increases by
16--24\% at $d=5$ (OPT: $0.0113\to0.0131$; LLaMA-1B:
$0.0112\to0.0139$), ruling out the interpretation that RAA failed to
redirect attention. This redistribution was not separately verified at
long distances ($d=40$--$90$) where behavioural nulls are reported; we
treat this as a limitation (\S\ref{sec:limitations}).

\subsection{Experiment 6: Multi-Fact Interference Probe}
\label{sec:exp6-method}

Each trial presents four fact sentences with fabricated rare names
sharing the same syntactic structure, drawn from eight distinct topic
sets to prevent results depending on a single pattern. Facts are
separated by WikiText filler text; the layout places two distractor
facts before the target, the target fact, a variable-length filler
gap, one distractor after the target, then the cue. The primary metric
is \emph{beats-distractors}: the percentage of trials where the target
name ranks higher than all three distractor names. 200 trials were run
per distance across $d \in \{10,20,\ldots,80\}$.

\paragraph{Methodological note.} Varying target distance
simultaneously varies inter-fact spacing; central conclusions rely on
cross-model comparisons at matched distance, where this confound is
held constant. With four candidates, random ranking yields
$\approx$25\% beats-distractors; models scoring substantially below
this threshold should be considered at floor on this task.

\section{Results}
\label{sec:results}

\subsection{Experiment 1: Baseline Attention Degradation}
\label{sec:exp1-results}

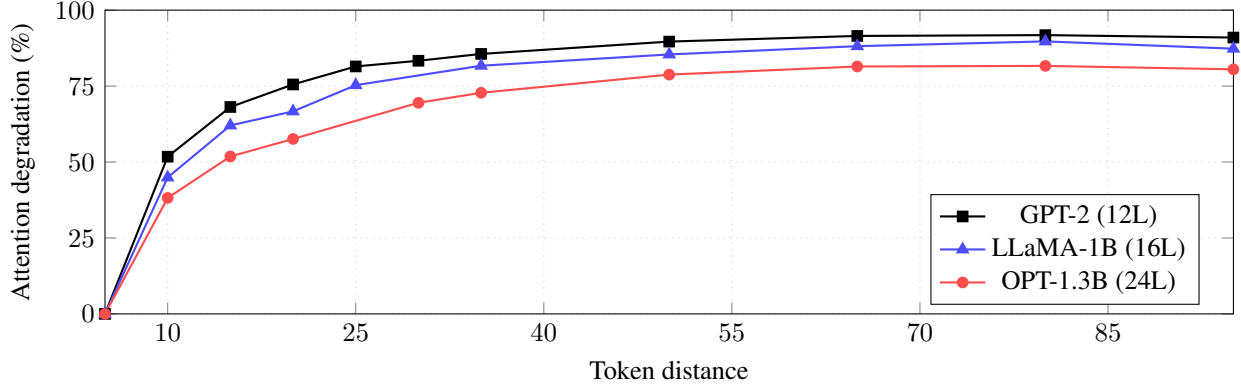
\begin{figure}[t]
\centering
\begin{tikzpicture}
\begin{axis}[
  width=\columnwidth,
  height=5.6cm,
  xlabel={Token distance},
  ylabel={Attention degradation (\%)},
  xmin=5, xmax=95,
  ymin=0, ymax=100,
  xtick={10,25,40,55,70,85},
  ytick={0,25,50,75,100},
  legend pos=south east,
  legend style={font=\small},
  grid=major,
  grid style={dotted, gray!40},
  tick label style={font=\small},
  label style={font=\small},
]
\addplot[color=black, mark=square*, mark size=1.8pt, thick]
  coordinates {
    (5,0) (10,51.76) (15,68.12) (20,75.54) (25,81.47) (30,83.34)
    (35,85.59) (50,89.65) (65,91.51) (80,91.78) (95,90.98)
  };
\addlegendentry{GPT-2 (12L)}
\addplot[color=blue!70, mark=triangle*, mark size=2.2pt, thick]
  coordinates {
    (5,0) (10,44.92) (15,62.05) (20,66.68) (25,75.31)
    (35,81.70) (50,85.42) (65,88.14) (80,89.71) (95,87.32)
  };
\addlegendentry{LLaMA-1B (16L)}
\addplot[color=red!70, mark=*, mark size=1.8pt, thick]
  coordinates {
    (5,0) (10,38.19) (15,51.83) (20,57.60) (30,69.50)
    (35,72.80) (50,78.78) (65,81.45) (80,81.65) (95,80.52)
  };
\addlegendentry{OPT-1.3B (24L)}
\end{axis}
\end{tikzpicture}
\caption{Attention degradation by token distance across architectures
(data from Appendix~\ref{app:exp1-tables}). All models follow an
exponential-then-plateau pattern; degradation rate is inversely
correlated with depth. All $d>10$ significant at $p<10^{-13}$.}
\label{fig:degradation}
\end{figure}

All models show statistically significant attention degradation
following a consistent exponential-then-plateau pattern
(Figure~\ref{fig:degradation}): a steep decline from $d=5$ to
approximately $d=40$--$50$, followed by an asymptotic plateau. All
distances beyond $d=10$ are significant at $p<10^{-13}$. The
degradation rate is inversely correlated with model depth: GPT-2
(12 layers) degrades fastest, LLaMA-1B (16 layers) at an intermediate
rate, and OPT-1.3B (24 layers) most gradually. LLaMA-3B follows the
same qualitative pattern, consistent with the depth-rate relationship.
Full per-distance values, including attention means and standard
deviations for each model, are provided in
Appendix~\ref{app:exp1-tables}.

\begin{table}[t]
\centering
\small
\begin{tabular}{lrrrr}
\toprule
Dist. & GPT-2 & LLaMA-1B & OPT-1.3B & Spread \\
\midrule
$d=10$ & 51.76\% & 44.92\% & 38.19\% & 13.57\,pp \\
$d=20$ & 75.54\% & 66.68\% & 57.60\% & 17.94\,pp \\
$d=35$ & 85.59\% & 81.70\% & 72.80\% & 12.79\,pp \\
$d=50$ & 89.65\% & 85.42\% & 78.78\% & 10.87\,pp \\
$d=65$ & 91.51\% & 88.14\% & 81.45\% & 10.06\,pp \\
$d=80$ & 91.78\% & 89.71\% & 81.65\% & 10.13\,pp \\
$d=95$ & 90.98\% & 87.32\% & 80.52\% & 10.46\,pp \\
\bottomrule
\end{tabular}
\caption{Cross-model attention degradation at selected distances.
pp = percentage points. Spread = GPT-2 minus OPT-1.3B. Measurements
span $d=5$--$95$; $d=100$ is excluded (\S\ref{sec:d100}).}
\label{tab:degradation}
\end{table}

Two further observations are noteworthy. First, the plateau levels
differ systematically: GPT-2 and LLaMA-1B plateau at 88--92\%, while
OPT-1.3B plateaus at 80--82\%, indicating that OPT retains more
absolute attention at long distances, not merely a slower approach to
the same asymptote. Second, absolute attention values at $d=80$
converge to the same order of magnitude across models
($\approx$0.0011--0.0021) despite very different baselines, far below
the $1/80 = 0.0125$ a uniform distribution would assign---consistent
with models concentrating residual long-range attention at specific
positions (cf.\ the attention sink, \citealt{xiao2023streamingllm})
rather than distributing it uniformly.

\subsubsection{The Distance-100 Anomaly}
\label{sec:d100}

All models show a pronounced anomaly at exactly $d=100$: GPT-2's
apparent degradation collapses to $+15.18\%$ (attention $=0.0156$),
while LLaMA-1B and OPT-1.3B show \emph{negative} degradation of
$-266.75\%$ and $-157.49\%$ respectively---attention far above the
$d=5$ baseline. This reflects the attention sink: when the target
position at distance 100 coincides with the beginning-of-sequence
position in the sampling scheme used, attention weights are
dramatically inflated. The distance is excluded from all substantive
analyses.

\subsubsection{Layer-wise Entropy Analysis}
\label{sec:entropy}

Attention entropy across layers reveals distinct architectural
signatures. \textbf{OPT-1.3B} shows monotonically decreasing entropy
across its 24 layers (3.91 nats at layer~1 to 0.79 nats at layer~23),
consistent with gradual, progressive refinement in which no single
layer resolves long-range dependencies alone. \textbf{LLaMA-1B} shows
a W-shaped profile: focused early (minimum 0.91 nats at layer~3),
diffuse in the middle (peak 2.25 nats at layer~8), refocused late
(1.02 nats at layer~14)---suggesting two-phase processing in which
early layers perform local syntactic integration and middle layers
broad contextual integration. \textbf{GPT-2} shows an oscillating
pattern with terminal diffusion (2.32 nats at layer~12); its diffuse
final-layer attention may explain its sensitivity to inserted
structural markers (\S\ref{sec:exp3-results}): when attention is not
strongly focused, any token that concentrates probability mass is
disproportionately influential.

\subsection{Experiment 2: Cloze Substitution}
\label{sec:exp2-results}

The cloze experiment reveals highly architecture-dependent patterns of
function token reliance, summarised in Table~\ref{tab:cloze-summary}.

\begin{table}[t]
\centering
\small
\begin{tabular}{llrr}
\toprule
Category & Model & Mean FN $\Delta p$ & Specificity \\
\midrule
Articles    & GPT-2     & $-0.022$ & 35\% (7/20) \\
Articles    & LLaMA-1B  & $-0.020$ & 38\% (8/21) \\
Articles    & OPT-1.3B  & $-0.054$ & 10\% (2/21) \\
\midrule
Prepositions & GPT-2    & $-0.028$ & 35\% (7/20) \\
Prepositions & LLaMA-1B & $-0.012$ &  0\% (0/21) \\
Prepositions & OPT-1.3B & $-0.088$ & 76\% (16/21) \\
\midrule
Punctuation & GPT-2     & $-0.009$ &  0\% (0/20) \\
Punctuation & LLaMA-1B  & $-0.009$ &  5\% (1/21) \\
Punctuation & OPT-1.3B  & $-0.026$ &  0\% (0/21) \\
\bottomrule
\end{tabular}
\caption{Cross-model cloze specificity (10 seeds per model). FN =
function token replaced with neutral token. Specificity = proportion
of distances where $|\text{FN}\,\Delta p| > |\text{Control}\,\Delta
p|$. GPT-2 results are reported over 20 distances; LLaMA-1B and
OPT-1.3B over 21, reflecting a one-token tokenizer boundary
difference. Inter-run $\Delta p$ estimates vary by up to $2\times$
across identical-configuration runs (\S\ref{sec:exp2-method}).}
\label{tab:cloze-summary}
\end{table}

\paragraph{OPT-1.3B: distance-dependent preposition specificity.}
Replacing prepositions causes probability drops increasing
approximately monotonically with distance: $-0.064$ at $d=5$,
deepening to $-0.078$ at $d=20$, $-0.088$ at $d=35$, and plateauing at
$\approx-0.095$ by $d=60$. The per-distance comparison
(Table~\ref{tab:opt-prep}) shows the specificity gap widening with
distance: at $d=5$ control damage actually exceeds preposition damage,
the crossover occurs at $\approx d=15$, and from $d=30$ onward
preposition damage consistently exceeds control damage by
$0.013$--$0.030$. After Bonferroni correction, specificity is
significant ($p<0.01$) at $d\in\{20,25,30,35,40\}$ and observed at 16
of 21 distances. This widening gap is the quantitative signature
predicted by the relay-chain hypothesis: as distance removes the
target from the range of direct attention, the absence of relay
anchors becomes proportionally more costly.

\begin{table}[t]
\centering
\small
\begin{tabular}{rrrc}
\toprule
Dist. & FN $\Delta p$ & Control $\Delta p$ & Specific? \\
\midrule
5   & $-0.064$ & $-0.126$ & --- \\
10  & $-0.073$ & $-0.076$ & --- \\
15  & $-0.076$ & $-0.071$ & \checkmark \\
20  & $-0.078$ & $-0.073$ & \checkmark \\
25  & $-0.078$ & $-0.081$ & --- \\
30  & $-0.087$ & $-0.071$ & \checkmark \\
35  & $-0.088$ & $-0.067$ & \checkmark \\
40  & $-0.090$ & $-0.072$ & \checkmark \\
45  & $-0.093$ & $-0.073$ & \checkmark \\
50  & $-0.094$ & $-0.063$ & \checkmark \\
55  & $-0.094$ & $-0.081$ & \checkmark \\
60  & $-0.095$ & $-0.068$ & \checkmark \\
65  & $-0.094$ & $-0.081$ & \checkmark \\
70  & $-0.095$ & $-0.082$ & \checkmark \\
75  & $-0.092$ & $-0.072$ & \checkmark \\
80  & $-0.092$ & $-0.073$ & \checkmark \\
85  & $-0.093$ & $-0.082$ & \checkmark \\
90  & $-0.093$ & $-0.063$ & \checkmark \\
95  & $-0.092$ & $-0.098$ & --- \\
105 & $-0.094$ & $-0.067$ & \checkmark \\
\bottomrule
\end{tabular}
\caption{OPT-1.3B per-distance preposition substitution results
(10 seeds, 100 samples). FN = prepositions replaced with neutral
token; Control = equal number of random content tokens replaced. The
anomalous distance $d=100$ is omitted (\S\ref{sec:d100}); specificity
is observed at 16 of the 21 measured distances.}
\label{tab:opt-prep}
\end{table}

In contrast, article and punctuation replacement in OPT causes
substantial probability drops ($-0.054$ and $-0.026$) but the control
condition consistently causes comparable or larger damage (specificity
10\% and 0\% respectively). Articles primarily signal definiteness---a
semantic property---whereas prepositions explicitly encode argument
structure; this difference in structural function maps directly onto
the difference in relay-chain specificity.

\paragraph{GPT-2: uniform non-specific dependence.}
Function token replacement causes consistent but remarkably flat
damage across distances ($-0.022$ articles, $-0.028$ prepositions,
$-0.009$ punctuation; cross-distance SD an order of magnitude smaller
than OPT's variation). Content tokens cause comparable damage, and
specificity is observed at only $\approx$35\% of distances. Weak
positive correlations between distance and probability change
(articles: $r=0.51$, $p=0.02$; punctuation: $r=0.80$, $p<0.001$)
indicate slightly \emph{less} damage at longer distances---the
opposite of the relay-chain prediction---though the absolute
magnitudes are too small to be practically meaningful. The relay-chain
hypothesis is not supported for GPT-2.

\paragraph{LLaMA-1B: reversal at long distances.}
Preposition replacement initially harms prediction ($-0.017$ at
$d=10$) but the effect diminishes monotonically to $-0.006$ by
$d=100$; rank-based measures show progressive \emph{improvement} when
prepositions are replaced, from $-19$ rank positions at $d=30$ to
$-372$ by $d=105$. This reversal is consistent with RoPE encoding
relative position directly in the attention computation, making
function tokens redundant as positional cues and potentially
introducing competing signals at long distances. Per-sample variance
is, however, very large (SD $>$ 600 rank positions), and none of the
LLaMA-1B cloze results survive Bonferroni correction; we report this
pattern as an exploratory observation that subsequent experiments
(\S\ref{sec:exp3-results}, \S\ref{sec:exp5-results}) independently
corroborate.

\subsection{Experiment 3: Function Token Insertion}
\label{sec:exp3-results}

The boundary strategy is consistently the most effective, with
strongest effects in the 40--80 token transition zone where attention
has partially but not fully degraded
(Table~\ref{tab:insertion}).

\begin{table}[t]
\centering
\small
\setlength{\tabcolsep}{3.5pt}
\begin{tabular}{rrrrrrr}
\toprule
 & \multicolumn{2}{c}{GPT-2} & \multicolumn{2}{c}{OPT-1.3B}
 & \multicolumn{2}{c}{LLaMA-1B} \\
\cmidrule(lr){2-3}\cmidrule(lr){4-5}\cmidrule(lr){6-7}
Dist. & $p$ & $\Delta p$ & $p$ & $\Delta p$ & $p$ & $\Delta p$ \\
\midrule
10 & 0.017* & $-0.009$ & 0.140 & $-0.015$ & 0.485 & $-0.010$ \\
20 & 0.046* & $-0.008$ & 0.020* & $-0.012$ & 0.054 & $-0.009$ \\
30 & 0.076  & $-0.005$ & 0.001* & $-0.011$ & 0.075 & $-0.006$ \\
40 & 0.009* & $-0.005$ & 0.025* & $-0.010$ & 0.042* & $-0.003$ \\
50 & 0.005* & $-0.006$ & 0.004* & $-0.010$ & 0.052 & $-0.000$ \\
60 & 0.004* & $-0.008$ & $<$0.001* & $-0.012$ & 0.048* & $-0.002$ \\
70 & 0.011* & $-0.008$ & 0.003* & $-0.011$ & 0.016* & $+0.003$ \\
80 & $<$0.001* & $-0.007$ & $<$0.001* & $-0.013$ & 0.039* & $+0.001$ \\
\bottomrule
\end{tabular}
\caption{Boundary insertion paired $t$-test results (function token
vs.\ content-token control). * = $p<0.05$. $\Delta p$ = mean
probability change for the function-token condition (negative = less
damage than baseline). 4--5 seeds per condition.}
\label{tab:insertion}
\end{table}

At $d=80$, comma insertion causes $4.1\times$ less probability damage
than random token insertion in GPT-2 ($-0.007$ vs.\ $-0.029$).
Significant distances in OPT-1.3B (20--80) converge with the distances
showing preposition specificity in Experiment~2, strengthening a
causal interpretation for this model. In LLaMA-1B, the probability
changes at $d=70$--$80$ turn slightly \emph{positive}, indicating
marginal harm from insertion at the longest distances---consistent
with the reversal in Experiment~2.

\paragraph{LLaMA-3.2-3B: rank-based evidence.}
LLaMA-3B's 128K-token vocabulary spreads probability mass so thinly
that per-token probability effects are on the order of $10^{-5}$ and
difficult to interpret; rank-based effects, however, are substantial
(Table~\ref{tab:llama3b}). Boundary insertion at $d=50$ improves the
target word's rank by $-3{,}294$ positions in the function-token
condition versus $-280$ for the control---a 7.5-fold difference---with
significant rank effects at $d=30$--$80$ ($p=0.014$--$0.045$).
Function token insertion thus genuinely reorders LLaMA-3B's
predictions even when probability changes are negligible.

\begin{table}[t]
\centering
\small
\setlength{\tabcolsep}{4pt}
\begin{tabular}{rrrr}
\toprule
Dist. & FN Rank $\Delta$ & Control Rank $\Delta$ & $p$ (rank) \\
\midrule
10 & $-507$   & $+500$   & 0.368 \\
20 & $-2354$  & $+602$   & 0.091 \\
30 & $-2364$  & $+317$   & 0.045* \\
40 & $-2813$  & $+835$   & 0.024* \\
50 & $-3294$  & $-280$   & 0.014* \\
60 & $-3853$  & $-978$   & 0.016* \\
70 & $-3567$  & $-434$   & 0.018* \\
80 & $-3746$  & $-1020$  & 0.025* \\
\bottomrule
\end{tabular}
\caption{LLaMA-3.2-3B boundary insertion, rank-based results (seeds
1--4). Rank $\Delta$ = mean change in target-word rank (negative =
improved prediction). * = $p<0.05$ (one-sample $t$-test vs.\ 0).}
\label{tab:llama3b}
\end{table}

\paragraph{Strategy comparison.}
The boundary strategy outperforms gap-fill and both interval
strategies across all models. Gap-fill helps in LLaMA-1B from $d=30$
onward but mostly causes harm in GPT-2 and OPT; interval-10 shows
significant rank improvements only in LLaMA-3B at $d=40$--$80$;
interval-20 is least effective overall. The superiority of
syntactically aligned insertion over density-matched uniform insertion
indicates that the benefit is tied to alignment with syntactic
structure, not to function-token density per se. Practically, this
means uniform-interval punctuation injection is a poor substitute for
even coarse clause-boundary heuristics.

\paragraph{Per-seed consistency.}
For GPT-2 at $d=40$--$80$, comma insertion beat the control in 4/5
seeds at $d=40$ and 5/5 seeds at $d=50$, $60$, and $80$; for OPT at
significant distances, 4--5 of 5 seeds; for LLaMA-3B, 4/4 completed
seeds at all distances. The significant $p$-values are therefore
driven by consistent cross-seed patterns rather than outlier seeds.

\subsection{Experiment 4: Relay Chain Coverage}
\label{sec:exp4-results}

\begin{table}[t]
\centering
\small
\begin{tabular}{llrrr}
\toprule
Category & Radius & GPT-2 & LLaMA-1B & OPT-1.3B \\
\midrule
Articles     & 60 tok. & 88.9\% & 88.4\% & 88.6\% \\
Prepositions & 25 tok. & 88.7\% & 88.2\% & 88.5\% \\
Punctuation  & 35 tok. & 86.4\% & 83.4\% & 84.8\% \\
\bottomrule
\end{tabular}
\caption{Relay chain coverage by function token category and
tokenizer. Coverage ranges from 83.4\% to 88.9\%; combined coverage
across all three categories is $\approx$96--98\%. Radii for articles
and punctuation derive from Experiment~2 conditions showing weak
specificity in most models; these figures primarily reflect corpus
density (\S\ref{sec:exp4-method}).}
\label{tab:coverage}
\end{table}

Coverage ranges from 83.4\% to 88.9\% across all category-tokenizer
combinations (Table~\ref{tab:coverage}), with $<$2\,pp difference
across tokenizers, indicating the analysis reflects a property of the
English text distribution rather than a tokenisation artefact. Gap
distributions are right-skewed with modal gaps of 2--5 tokens and
tails to 40--50 tokens. The three categories are complementary:
positions uncovered by one category are typically covered by another,
raising combined coverage to $\approx$96--98\%. Uncovered positions
concentrate in long spans of uninterrupted content words (multi-modifier
noun phrases, dense technical passages, numerical sequences), which
are comparatively rare in encyclopaedic prose.

\subsection{Experiment 5: Relay-Aware Attention --- The Central Causal
Test}
\label{sec:exp5-results}

\begin{table}[t]
\centering
\small
\setlength{\tabcolsep}{3.5pt}
\begin{tabular}{lrllrr}
\toprule
Model & L & Enc. & Effect & Best & Worst \\
\midrule
GPT-2    & 12 & Abs. & Null     & $+0.33$\,pp         & ---                  \\
OPT-1.3B & 24 & Abs. & Mixed    & $+2.82$\,pp$^\dagger$ & $-1.53$\,pp$^\dagger$ \\
LLaMA-1B & 16 & RoPE & Null     & $+0.68$\,pp         & $-0.95$\,pp          \\
LLaMA-3B & 28 & RoPE & Neg.$^*$ & ---                 & $-5.03$\,pp$^\dagger$ \\
\bottomrule
\end{tabular}
\caption{RAA effects across architectures. L=layers. Best/Worst =
largest observed improvement or harm at any distance for the
strongest-effect $\delta$. $\dagger$ = significant at $p<0.05$. GPT-2
and LLaMA-1B show no significant improvement at any distance. The
LLaMA-3B result is based on 2 seeds and should be treated as
indicative only ($^*$).}
\label{tab:raa}
\end{table}

\paragraph{GPT-2 and LLaMA-1B: null effects.} No significant
improvement at any distance survives Bonferroni correction.

\paragraph{LLaMA-3B: preliminary negative effect.} All $\delta$ values
increase degradation ($-5.03$\,pp at $d=10$, $\delta=0.1$), extending
the reversal pattern from Experiments~2--3. This result rests on 2
seeds and should be treated as a preliminary indication of harm rather
than a robust finding.

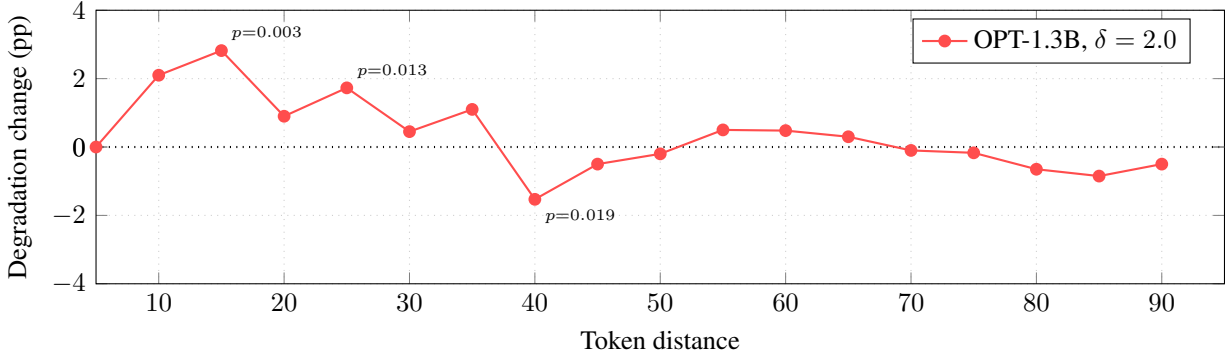
\begin{figure}[t]
\centering
\begin{tikzpicture}
\begin{axis}[
  width=\columnwidth,
  height=5.2cm,
  xlabel={Token distance},
  ylabel={Degradation change (pp)},
  xmin=5, xmax=95,
  ymin=-4, ymax=4,
  xtick={10,20,30,40,50,60,70,80,90},
  legend pos=north east,
  legend style={font=\small},
  grid=major,
  grid style={dotted, gray!40},
  tick label style={font=\small},
  label style={font=\small},
  extra y ticks={0},
  extra y tick style={grid=major, grid style={black, thick}},
]
\addplot[color=red!70, mark=*, mark size=2.0pt, thick]
  coordinates {
    (5,  0.00)
    (10, 2.10)
    (15, 2.82)
    (20, 0.90)
    (25, 1.73)
    (30, 0.45)
    (35, 1.10)
    (40,-1.53)
    (45,-0.50)
    (50,-0.20)
    (55, 0.50)
    (60, 0.48)
    (65, 0.30)
    (70,-0.10)
    (75,-0.17)
    (80,-0.65)
    (85,-0.85)
    (90,-0.50)
  };
\addlegendentry{OPT-1.3B, $\delta=2.0$}
\node[font=\tiny, above right] at (axis cs:15,2.82) {$p{=}0.003$};
\node[font=\tiny, above right] at (axis cs:25,1.73) {$p{=}0.013$};
\node[font=\tiny, below right] at (axis cs:40,-1.53) {$p{=}0.019$};
\end{axis}
\end{tikzpicture}
\caption{RAA effect on OPT-1.3B at $\delta=2.0$ (positive =
improvement, negative = harm). Improvements at short distances are
offset by harm at longer distances; the net effect is approximately
zero.}
\label{fig:raa}
\end{figure}

\paragraph{OPT-1.3B: distance-dependent mixed effect.} Short distances
($d=10$--$30$) show mean $+1.6$\,pp improvement (significant at
$d=15$: $+2.82$\,pp, $p=0.003$; $d=25$: $+1.73$\,pp, $p=0.013$). Long
distances ($d=40$--$90$) show mean $-0.3$\,pp harm (significant at
$d=40$: $-1.53$\,pp, $p=0.019$). The net effect across all distances
is approximately zero: RAA shifts \emph{where} degradation occurs
without reducing its overall magnitude (Figure~\ref{fig:raa}).

This is the central finding: a sound intervention that verifiably
increases attention to function token positions by 16--24\% does not
produce the improvements that the relay-chain interpretation of mean
attention would predict in three of four models. Even in the most
favourable case (OPT-1.3B)---the model whose correlational profile
most strongly supports the relay chain---the redistribution nets to
approximately zero.

\subsection{Experiment 6: Multi-Fact Interference Probe}
\label{sec:exp6-results}

\begin{table}[t]
\centering
\small
\begin{tabular}{rrrrr}
\toprule
Dist. & distilgpt2 & GPT-2 & LLaMA-1B & OPT-1.3B \\
\midrule
$d=10$ & 17.5\% & 14.5\% & 35.5\% & 36.0\% \\
$d=30$ & 16.0\% & 14.5\% & 40.0\% & 46.0\% \\
$d=50$ & 13.0\% & 15.5\% & 44.0\% & 67.0\% \\
$d=70$ & 16.0\% & 17.0\% & 50.0\% & 76.0\% \\
$d=80$ & 13.0\% & 10.0\% & 58.5\% & 84.0\% \\
\bottomrule
\end{tabular}
\caption{Beats-distractors \% across models and distances (200 trials
per distance). Chance level $\approx$25\%; the GPT-2 family scores
below chance at all distances.}
\label{tab:retrieval}
\end{table}

\paragraph{distilgpt2 vs.\ GPT-2.} Paired $t$-tests yield $p=0.41$
(beats-distractors, primary metric) and $p=0.44$ (chose-distractor)
across eight distance points. However, both models score 10--17.5\%
beats-distractors at every distance---below the $\approx$25\% chance
level for four candidates---indicating both are at floor on this task.
Statistical indistinguishability at floor is weaker evidence than a
mid-range comparison, and we caution against over-interpreting this
comparison.

\paragraph{LLaMA-1B vs.\ OPT-1.3B: the stronger comparison.} Both
models substantially outperform the GPT-2 family. At $d=10$,
performance is nearly identical (35.5\% vs.\ 36.0\%); OPT's advantage
then grows markedly with distance, reaching 84.0\% vs.\ 58.5\% at
$d=80$---despite LLaMA-1B having a \emph{steeper} mean attention
degradation curve (Table~\ref{tab:degradation}). This divergence, in
the direction opposite to what degradation rate would predict,
provides the most compelling evidence that mean degradation rate does
not determine retrieval capacity. The dividing line is model capacity
($\sim$1.2--1.3B parameters), not degradation rate.

\section{Discussion}
\label{sec:discussion}

\subsection{The Architecture-Dependent Relay Chain}

The correlational and interventional evidence (Experiments~1--4)
establishes that the function token relay-chain mechanism is real but
architecture-dependent, interpretable through three interacting
factors: positional encoding scheme, depth, and entropy profile.

\textbf{OPT-1.3B} is most consistent with the relay-chain
interpretation. Its absolute positional encoding does not encode
relative distance in the attention computation, so the model must
learn to use token content as a positional substitute at long
distances; its 24-layer depth and monotonically decreasing entropy
profile create the conditions for function tokens to serve as
persistent anchors. The convergence of Experiments~2 and~3---specific
preposition dependence and significant boundary insertion effects at
the same distance ranges---provides compelling evidence for a causal
rather than merely correlational mechanism in this model.

\textbf{GPT-2} presents an apparent paradox: strong boundary insertion
effects (Experiment~3, $p<0.001$ at $d=80$) but no cloze specificity
(Experiment~2). This resolves by recognising that Experiment~2 tests
\emph{removing} existing function tokens while Experiment~3 tests
\emph{adding} structural tokens. GPT-2 can utilise structural signals
when provided but does not route attention specifically through
existing function tokens; in its shallow, capacity-limited
architecture, content tokens carry as much predictive information as
function tokens.

\textbf{LLaMA} provides the most theoretically informative results.
LLaMA-1B's monotonic reversal with distance is consistent with RoPE
encoding relative position directly via query-key rotations: even
without structural tokens at intermediate positions, attention scores
already contain relative-distance information, making function tokens
redundant in the positional dimension and---at long
distances---potentially a source of attention dilution. LLaMA-3B's
RAA result extends this: function token emphasis is actively harmful
for RoPE models (though this rests on 2 seeds and warrants
replication).

This architectural contingency yields a testable prediction: models
with absolute positional encoding and greater depth should show
stronger function token dependence than shallow or RoPE-based models,
all else being equal.

\subsection{Mean Attention Is Largely Descriptive, Not Prescriptive}
\label{sec:descriptive}

Experiments~5 and~6 jointly establish that mean cross-positional
attention degradation is at best a weak, architecture-specific
contributor to contextual processing. This conclusion extends the
attention-as-explanation debate \citep{jain2019attention,
wiegreffe2019attention, bibal2022attention} from encoder classifiers
to autoregressive LMs, and from correlation-based faithfulness
analysis to direct intervention: we verify that attention mass moves
before asking whether behaviour changes.

The RAA experiments verify a 16--24\% redistribution of attention mass
without improvement in three of four models; even in OPT-1.3B the net
effect is approximately zero. The retrieval probes show that
degradation rate does not predict task performance, with the
LLaMA-1B/OPT-1.3B divergence running opposite to the degradation-rate
prediction. This is exactly the pattern expected if mean attention is
a descriptive summary of a downstream computation rather than the
computational variable itself.

\subsection{Resolving the Comma Insertion Paradox}
\label{sec:comma-paradox}

Experiment~3 shows comma insertion reduces prediction degradation
(causal benefit), while Experiment~5 shows biasing attention toward
\emph{existing} function tokens does not help. The resolution: comma
insertion adds a \emph{new} token to the sequence, creating a fresh
computational node whose hidden state is shaped by the full forward
pass. RAA merely increases attention to positions whose
representations are already fixed. The benefit of function tokens lies
in what they compute---the contextual information accumulated in their
hidden states---not in how much attention they receive. This is
consistent with \citet{zhang2025function}'s proposal that function
tokens activate predictive features, and with
\citet{meng2022locating}: factual associations live in MLP layers, not
attention heads.

\subsection{Theoretical Contributions}

Beyond the specific empirical findings, this work makes two broader
theoretical contributions. First, it introduces
\emph{architecture-conditioned function token dependence}: the degree
to which a model relies on function tokens as relay nodes is a
predictable function of its positional encoding scheme and depth,
providing a framework for anticipating model behaviour under
function-token manipulation without per-model experimentation. Second,
it establishes the short-term window (5--100 tokens) as a distinct
degradation regime, parameterisable by initial decay rate (varying
with depth) and plateau level (varying with depth and encoding),
separate from the long-context regime studied in prior work.

\subsection{Practical Implications}
\label{sec:practical}

\paragraph{Prompt engineering.} Strategic comma placement at clause
boundaries reduces prediction degradation in the 40--80 token range
for the GPT-2- and OPT-family \emph{base} models tested. For
RoPE-based models (LLaMA), the benefit is less consistent and may be
negative at long distances. Conversely, compression strategies that
strip articles, prepositions, or punctuation to save tokens may
degrade relay-chain coverage for absolute-encoding models. These
findings derive from base pretrained models below 3.5B parameters and
should not be extrapolated to instruction-tuned or frontier-scale
models without further investigation.

\paragraph{Retrieval-augmented generation.} Natural function token
density in standard English prose suffices for corpus-level coverage,
but heavily extractive or telegraphic content---product fragments,
JSON, code---lacks this density and may be more susceptible to
contextual degradation; restoring function token density before
context insertion is a candidate preprocessing step.

\paragraph{Interpretability methodology.} Studies using mean attention
weights as evidence for contextual processing may be measuring a
correlate rather than a cause. The correlation-to-intervention gap
documented here suggests head-level analysis, activation patching
\citep{meng2022locating}, or residual stream decomposition provide
more causally informative views than aggregate attention statistics.

\paragraph{KV-cache eviction.} Our findings carry a caution for
inference-efficiency methods that select or evict cached tokens based
on accumulated attention scores, such as H2O \citep{zhang2023h2o} and
SnapKV \citep{li2024snapkv}. These methods implicitly assume that
attention mass to a position tracks that position's causal
contribution to downstream prediction---precisely the assumption our
intervention tests. The dissociation we observe between attention mass
and behavioural contribution, together with the comma-paradox
resolution (\S\ref{sec:comma-paradox}), suggests that a token's value
lies in what its hidden state has computed rather than in how much
attention it subsequently receives; eviction policies keyed to
attention scores may therefore discard tokens whose representations
remain causally important. Evaluating representation-aware eviction
criteria against attention-score baselines is a natural follow-up.

\paragraph{Architecture design and evaluation.} Naive attention-bias
toward structural tokens is not the right operationalisation of the
relay chain; layer-specific biasing or representation-level
interventions targeting what function tokens \emph{compute} are more
promising. The degradation curve of Experiment~1 also offers a
task-independent diagnostic of effective contextual reach that could
complement standard benchmarks: models with similar benchmark scores
but different degradation profiles may fail in different contexts.

\subsection{Relationship to Mechanistic Interpretability}

The head-specialisation findings of \citet{clark2019bert} and
\citet{voita2019analyzing} suggest specific heads may track function
token positions. Our contribution is behavioural and
architecture-comparative: it establishes that the function token
mechanism exists, characterises its architecture-dependence, and shows
that its causal pathway is representational rather than
attention-routing. A natural next step is to use our perturbations
(substitution, insertion, logit biasing) as causal probes within an
activation-patching framework to localise the responsible components.

\subsection{Generalisability}

WikiText is formal English encyclopaedic prose; function token density
is likely lower in conversational, telegraphic, or code-heavy text and
the relay chain correspondingly sparser. The function token hypothesis
is implicitly language-specific: agglutinative languages encoding
grammatical relations morphologically may show fundamentally different
relay properties, making cross-linguistic replication a priority. All
models are base pretrained models below 3.5B parameters;
instruction-tuned or RLHF-aligned variants are known to alter
attention distributions, and whether alignment strengthens or weakens
the relay-chain mechanism is an open empirical question.

\section{Limitations}
\label{sec:limitations}

\paragraph{Statistical power.} Experiment~3 uses 4--5 seeds (3--4 df);
LLaMA RAA experiments used 2 complete seeds due to computational
constraints (consumer Apple Silicon hardware). The convergent pattern
across distances and conditions provides more robust support than any
single test, but the LLaMA-3B negative result in particular should be
treated as preliminary; a properly powered replication with $\geq$5
seeds is needed.

\paragraph{RAA verification at long distances.} Attention
redistribution was verified at $d=5$ only. At the long distances where
behavioural nulls matter most ($d=40$--$90$), softmax saturation or
attention-sink absorption could in principle have absorbed the bias,
making the null ambiguous at those distances.

\paragraph{Multiple comparisons.} Bonferroni correction across
$\approx$63 tests in Experiment~2 is conservative; Benjamini-Hochberg
FDR control would be more appropriate for this exploratory design and
would likely identify additional significant effects.

\paragraph{Substitution token confound.} The newline substitution
token is itself structurally meaningful in some tokenizers; a rare
out-of-vocabulary token would be cleaner.

\paragraph{Control position matching.} Experiment~3's control inserts
at random positions rather than at the boundary positions used by the
function-token strategies, so observed differences partially reflect
insertion position as well as token identity; position-matched
controls would isolate type-specific effects.

\paragraph{Dataset, language, and model scope.} All experiments use
English WikiText and base pretrained models of 82M--3.21B parameters
truncated to 512-token sequences. Results may not generalise to other
registers, languages, instruction-tuned variants, larger models, or
longer sequence regimes (1K--128K tokens).

\paragraph{Attention metric.} Mean attention across all layers and
heads masks head-level heterogeneity; syntactically specialised heads
may show different distance-dependence than the average.

\paragraph{Experiment 4 effective radii.} Radii for articles and
punctuation derive from Experiment~2 conditions showing weak or absent
specificity in most models; coverage figures for these categories
primarily reflect corpus token density rather than model-specific
anchoring.

\paragraph{RAA scope.} RAA is one operationalisation of the relay
chain, using uniform logit biasing across all layers and heads.
Layer-specific biasing, head-specific biasing, and value-pathway
interventions remain unexplored.

\paragraph{Softmax normalisation artefact.} Mean attention to a
position $d$ tokens back must decline with $d$ to some degree because
softmax mass is conserved over a growing context; part of the
exponential-then-plateau shape is therefore arithmetic rather than
mechanistic---which independently supports the interpretation of mean
attention degradation as a descriptive summary statistic rather than a
causal variable.

\section{Conclusion}
\label{sec:conclusion}

We presented six coordinated experiments investigating short-term
attention degradation, function token anchoring, and the causal status
of mean cross-positional attention across transformer language models.
The function token relay-chain mechanism is real but
architecture-dependent: OPT-1.3B shows distance-dependent preposition
specificity and responds to boundary insertion; GPT-2 shows
non-specific uniform dependence but benefits from inserted structural
tokens; LLaMA shows reversal effects consistent with redundant
positional information encoded by RoPE.

Attempts to amplify this mechanism through Relay-Aware Attention
reveal its limits: null effects on GPT-2 and LLaMA-1B, preliminary
indication of active harm on LLaMA-3B, and a small distance-dependent
mixed effect on OPT-1.3B that nets to approximately zero, despite
verified 16--24\% attention redistribution. Multi-fact interference
probes confirm that mean degradation rate does not predict retrieval
accuracy, with the LLaMA-1B vs.\ OPT-1.3B comparison providing the
strongest evidence.

These findings establish that mean cross-positional attention
degradation is largely descriptive rather than prescriptive. Where
function tokens contribute to contextual processing, they do so
primarily through token representations---the contextual information
accumulated in their hidden states---rather than through the attention
routing that mean cross-positional analysis captures. This refines the
function token hypothesis from a universal claim to an
architecture-conditioned mechanism, extends the
attention-as-explanation debate to autoregressive models with a
verified interventional design, and cautions against treating
aggregate attention statistics---in interpretability analyses or in
attention-score-based inference optimisations---as proxies for causal
importance.

\appendix

\section{Full Experiment 1 Degradation Tables}
\label{app:exp1-tables}

\begin{table}[h]
\centering
\small
\setlength{\tabcolsep}{4pt}
\begin{tabular}{rrrrr}
\toprule
Dist. & Attn Mean & Attn SD & Degrad.\ \% & $p$ \\
\midrule
5  & 0.018373 & 0.000172 & 0.00 (base) & --- \\
10 & 0.008863 & 0.000109 & 51.76 & 5.1e-17 \\
15 & 0.005858 & 0.000077 & 68.12 & 1.1e-17 \\
20 & 0.004493 & 0.000045 & 75.54 & 4.2e-18 \\
25 & 0.003404 & 0.000032 & 81.47 & 8.8e-19 \\
30 & 0.003062 & 0.000050 & 83.34 & 2.7e-19 \\
35 & 0.002647 & 0.000045 & 85.59 & 1.4e-18 \\
50 & 0.001902 & 0.000031 & 89.65 & 7.9e-19 \\
65 & 0.001560 & 0.000022 & 91.51 & 2.0e-19 \\
80 & 0.001510 & 0.000022 & 91.78 & 2.4e-19 \\
95 & 0.001658 & 0.000037 & 90.98 & 1.9e-19 \\
\bottomrule
\end{tabular}
\caption{GPT-2 attention degradation (10 seeds, 100 samples). Selected
distances shown; all $d \geq 10$ significant at $p<5\times10^{-17}$.}
\label{tab:gpt2-full}
\end{table}

\begin{table}[h]
\centering
\small
\setlength{\tabcolsep}{4pt}
\begin{tabular}{rrrrr}
\toprule
Dist. & Attn Mean & Attn SD & Degrad.\ \% & $p$ \\
\midrule
5  & 0.010995 & 0.000131 & 0.00 (base) & --- \\
10 & 0.006056 & 0.000062 & 44.92 & 4.5e-15 \\
15 & 0.004172 & 0.000073 & 62.05 & 2.5e-16 \\
20 & 0.003664 & 0.000083 & 66.68 & 4.3e-16 \\
25 & 0.002715 & 0.000051 & 75.31 & 1.7e-17 \\
35 & 0.002013 & 0.000056 & 81.70 & 4.0e-17 \\
50 & 0.001604 & 0.000042 & 85.42 & 3.6e-18 \\
65 & 0.001304 & 0.000032 & 88.14 & 4.6e-18 \\
80 & 0.001132 & 0.000033 & 89.71 & 3.0e-18 \\
95 & 0.001394 & 0.000028 & 87.32 & 2.5e-18 \\
\bottomrule
\end{tabular}
\caption{LLaMA-3.2-1B attention degradation (10 seeds, 100 samples).}
\label{tab:llama-full}
\end{table}

\begin{table}[h]
\centering
\small
\setlength{\tabcolsep}{4pt}
\begin{tabular}{rrrrr}
\toprule
Dist. & Attn Mean & Attn SD & Degrad.\ \% & $p$ \\
\midrule
5  & 0.011293 & 0.000115 & 0.00 (base) & --- \\
10 & 0.006979 & 0.000094 & 38.19 & 1.5e-14 \\
15 & 0.005440 & 0.000076 & 51.83 & 7.0e-16 \\
20 & 0.004788 & 0.000053 & 57.60 & 2.0e-16 \\
30 & 0.003445 & 0.000042 & 69.50 & 2.1e-17 \\
35 & 0.003072 & 0.000031 & 72.80 & 4.4e-18 \\
50 & 0.002396 & 0.000031 & 78.78 & 5.5e-18 \\
65 & 0.002095 & 0.000023 & 81.45 & 4.8e-18 \\
80 & 0.002072 & 0.000017 & 81.65 & 1.7e-18 \\
95 & 0.002200 & 0.000038 & 80.52 & 7.6e-18 \\
\bottomrule
\end{tabular}
\caption{OPT-1.3B attention degradation (10 seeds, 100 samples).}
\label{tab:opt-full}
\end{table}

\section*{Code and Data Availability}
Code, per-experiment configuration files, and result tables supporting
the findings of this paper are available from the authors on
reasonable request.

\section*{Acknowledgements}
This work derives from the first author's Level 7 project at
London Metropolitan University / Islington College, supervised by the
second author.

\bibliographystyle{plainnat}
\bibliography{references_full}

\end{document}